\documentclass[conference]{IEEEtran}
\usepackage{float}

\usepackage[utf8]{inputenc}
\usepackage[T1]{fontenc}
\usepackage[swedish,english]{babel}
\usepackage{amsfonts}
\usepackage{amsmath}

\usepackage{graphicx, float, subfigure, blindtext}

\makeatletter
\newcommand{\subalign}[1]{%
  \vcenter{%
    \Let@ \restore@math@cr \default@tag
    \baselineskip\fontdimen10 \scriptfont\tw@
    \advance\baselineskip\fontdimen12 \scriptfont\tw@
    \lineskip\thr@@\fontdimen8 \scriptfont\thr@@
    \lineskiplimit\lineskip
    \ialign{\hfil$\m@th\scriptstyle##$&$\m@th\scriptstyle{}##$\hfil\crcr
      #1\crcr
    }%
  }%
}
\makeatother

\DeclareMathOperator*{\argmax}{arg\,max}

\newcommand\IEEEhyperrefsetup{
bookmarks=true,bookmarksnumbered=true,%
colorlinks=true,linkcolor={black},citecolor={black},urlcolor={black}%
}

\usepackage[\IEEEhyperrefsetup, pdftex]{hyperref}

\usepackage{mathtools}

\usepackage[nolist,nohyperlinks]{acronym}
\usepackage{cleveref}
\graphicspath{{images/}}
\acrodef{IC}[IC]{Integrated Circuit}
\newacro{svm}[SVM]{Support Vector Machine}

\author{\IEEEauthorblockN{
Henry Charlesworth\IEEEauthorrefmark{1},
Adrian Millea\IEEEauthorrefmark{1}, 
Eddie Pottrill\IEEEauthorrefmark{2},
Rich Riley\IEEEauthorrefmark{1}
}

\IEEEauthorblockA{
\IEEEauthorrefmark{1}Rowden Technologies, Bristol, UK \\
\IEEEauthorrefmark{2} Defence Science and Technology Laboratory, UK
}}

\title{Room Clearance with Feudal Hierarchical Reinforcement Learning}
\begin{document}
\maketitle
\begin{abstract}
Reinforcement learning (RL) is a general framework that allows systems to learn autonomously through trial-and-error interaction with their environment. In recent years combining RL with expressive, high-capacity neural network models has led to impressive performance in a diverse range of domains. However, dealing with the large state and action spaces often required for problems in the real world still remains a significant challenge. In this paper we introduce a new simulation environment, ``Gambit", designed as a tool to build scenarios that can drive RL research in a direction useful for military analysis. Using this environment we focus on an abstracted and simplified room clearance scenario, where a team of blue agents have to make their way through a building and ensure that all rooms are cleared of (and remain clear) of enemy red agents. We implement a multi-agent version of feudal hierarchical RL that introduces a command hierarchy where a commander at the higher level sends orders to multiple agents at the lower level who simply have to learn to follow these orders. We find that breaking the task down in this way allows us to solve a number of non-trivial floorplans that require the coordination of multiple agents much more efficiently than the standard baseline RL algorithms we compare with. We then go on to explore how qualitatively different behaviour can emerge depending on what we prioritise in the agent's reward function (e.g. clearing the building quickly vs. prioritising rescuing civilians).
\end{abstract}
\begin{center}
\end{center}
\section{Introduction}
Reinforcement learning (RL) provides a general framework for autonomously learning to make decisions in order to maximise a cumulative sum of rewards. In this paper we introduce a new simulation environment designed as a simple testbed for demonstrating the utility of RL as a tool for concept analysis with military applications as well as to aid with the development of research in the field of multi-agent RL. RL can be applied to a wide range of tasks, and in recent years has found significant success in solving board games such as Chess \cite{alphazero} and Go \cite{alphago}, video games such as Starcraft 2 \cite{alphastarblog} and DOTA 2 \cite{openai2019dota}, and a variety of robotic control problems \cite{visuomotor, openai2021asymmetric}. However when looking to scale RL to complex problems of interest in the real world a number of challenges remain. If the state and action spaces are simply considered as huge unstructured spaces to search through then complex problems can easily become intractable for standard RL algorithms --- particularly if the temporal resolution is high and a very large number of decisions have to be made in order to complete a task. These problems can be further exacerbated in situations where multiple agents are learning concurrently since this can introduce non-stationarity into the environment (from each individual's point of view the environment includes the behaviour of all other agents which is constantly changing as they learn). Our goal was to create an environment which is simple enough to be accessible with existing RL algorithms, but with enough complexity to both a) generate interesting behaviour and strategies, and b) allow for research into new RL algorithms that can lead to improved performance. 

One promising framework for addressing some of the aforementioned issues is Hierarchical reinforcement learning (HRL) \cite{HRL}, which aims to break down the learning process into more manageable sub-tasks at differing levels of granularity. The basic idea is to create a hierarchy, where simpler sub-tasks are learned at the lower-levels, whilst higher levels of the hierarchy learn to control when these sub-tasks are initiated. This allows the higher levels to work at a greater level of abstraction, as well as at a lower temporal resolution such that fewer decisions have to be made. In principle this should lead to substantially more efficient learning. Another potential benefit is that the sub-tasks learned by the lower-levels of the hierarchy can be re-useable for solving new downstream tasks without needing to restart the whole training process from scratch.

We use the proposed environment to apply a particular form of HRL – known as “Feudal HRL” \cite{feudalrl92} – to scenarios that involve the coordination and cooperation of multiple agents. The Feudal HRL framework introduces a command hierarchy, where agents at a higher level (“commanders”) give agents at the level below orders to follow (who could potentially have agents at another level below them, to which they in turn also give orders). Crucially, agents at each level of the hierarchy can view the environment at different resolutions, appropriate for the kinds of task they are given, and only ever receive rewards from their commander when they successfully complete an order. This ``reward hiding" means that only the commander(s) at the top of the hierarchy ever see the reward provided by the environment, allowing the sub-commanders at lower-levels to focus completely on learning how to carry out the orders they are given.

One of the key issues that arises in multi-agent reinforcement learning is that the environment, from the point of view of an individual agent, is non-stationary. This is because, assuming that the agents interact in some way, the states/rewards that an agent sees depends on the policies followed by the other agents which are constantly changing as they learn. Whilst Feudal HRL is not as general as other multi-agent RL algorithms due to the imposed chain of command, in situations where it is applicable it has the potential to lead to significantly more stable training. This is because, from a given sub-commander’s perspective, once the agents below them have learned how to correctly follow their orders then the environment appears substantially more stationary. In particular, if there is a single commander at the top then once all the agents at the lower-levels of the hierarchy have learned how to follow their orders then the environment is entirely stationary from the commander's perspective.

\begin{figure*}
    \centering
    \includegraphics[width=1.0\linewidth]{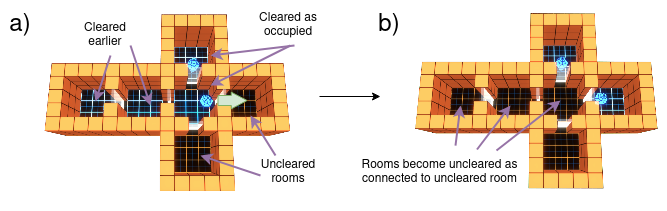}
    \caption{Simple example of the room clearance problem. In a), four rooms are ``cleared", however in b) the agent leaves a cleared room which is still connected to a room which is currently unclear. This immediately propagates to all connected unoccupied rooms.}
    \label{fig:example_roomclearance}
\end{figure*}

The scenario we focus on is that of “room clearance”, where a building consisting of a number of rooms is being stormed by “soldiers” (blue agents) and explored in order to ensure that every room is and remains cleared of enemy combatants (red agents). Example scenarios can also include “civilians” (white agents) which may need to be rescued. We show that Feudal HRL, where a commander issues orders to each of the blue agents, is able to solve a range of interesting, non-trivial scenarios in an efficient manner. Additionally, we explore how employing different reward functions can lead to a variety of different strategies being learned for a given scenario (e.g. prioritising rescuing civilians over clearing the building quickly), demonstrating how the strategies which are learned can vary according to specified priorities.

\section{Related Work}
Dayan \& Hinton first proposed the idea of introducing a command hierarchy into the RL framework \cite{feudalrl92}. Here they emphasised the idea of reward and information hiding, such that the lower level of the hierarchy can focus entirely on completing the orders given and only has access to information at the relevant level of granularity. In a simple proof of concept they demonstrated that a tabular Q-learning implementation using these ideas solved a basic maze problem with substantially improved efficiency compared to a standard Q-learning approach. Vezhnevets et al. \cite{vezhnevets2017feudal} extended this work into the deep RL domain, combining the ideas of this framework with deep Q-networks \cite{dqn}. This allowed them to scale the approach to significantly more challenging environments and demonstrate that it could outperform strong baselines on problems that required long-term credit assignment. There are two key differences compared to the approach we take --- firstly, their ``manager" agent learns the abstract goals it sets the ``worker" agent, whereas for us the set of goals the commander can issue is fixed. Secondly, they only consider a situation with a single manager and a single worker, whilst we consider multiple workers. Indeed, algorithmically our approach is most similar to Ahilan and Dayan \cite{ahilan2019feudal}, who extend the feudal RL framework to the cooperative multi-agent setting. They test this approach on a number of simple environments that require cooperation and communication between agents and demonstrate that it offers considerable benefits over standard multi-agent RL algorithms.

\section{Environment}
We introduce a simple but flexible gridworld environment that supports a variety of different rulesets, as well as including a scenario editor that easily allows us to create buildings with fully customisable floorplans and different initial conditions for the agents, civilians and enemy combatants. Whilst other rulesets are supported (e.g. capture the flag), in this paper we focus on the ``room clearance" ruleset. The basic set up is that a commander issues orders to a group of agents with the aim of ensuring that the building is safe and clear of enemy combatants. A room is considered to be ``clear" either if it is occupied by an agent with no enemy combatants present, or if it has already been cleared and all the rooms it is connected to are also clear. As such, when a room is uncleared this property propagates throughout the building to all connected rooms that are unoccupied at that point (Figure \ref{fig:example_roomclearance} shows a simple example of how this can happen). This means that the commander has to plan and carefully coordinate the movement of the agents to ensure that rooms are not unintentionally made unclear. To actually perform the calculation to determine whether rooms are clear or not we use the adjacency matrix of the room graph, $\mathbf{A} \in \mathbb{R}^{m \times m}$, where $m$ is the number of rooms and $A_{ij} = 1$ if room i and j are connected and 0 otherwise. Let us define a vector $\mathbf{u} \in \mathbb{R}^{m \times 1}$, such that $u_i = 1$ if room $i$ is unclear and otherwise $u_i=0$, and a vector $\mathbf{v} \in \mathbb{R}^{m \times 1}$, such that $v_i = 1$ if room $i$ is unoccupied by an agent and otherwise $v_i=0$. If we consider $\mathbf{u}$ as a signal, we can think of the action of multiplying $\mathbf{u}$ by $\mathbf{A}$ as effectively propagating the signal into directly connected rooms (here propagating the signal that a room is uncleared into neighbouring rooms). We can then perform element-wise multiplication ($\odot$) between the result and the vector of unoccupied rooms to ensure that this signal is only propagated to rooms that do not contain an agent, and clip the result such that the values are always binary. We then repeatedly apply this process:

\begin{equation}
    \mathbf{u} = \text{clip}\left((\mathbf{A u}) \odot \mathbf{v}, 0, 1\right)
\end{equation}
until $\mathbf{u}$ stops changing, which is the point at which the ``unclear" signal has been successfully propagated to all unoccupied, connected rooms.

In the environment, the commander is not a physical entity but simply issues commands to each individual agent whenever necessary. In the simplest set up the set of available commands is equal to the number of doors in the room that the agent receiving the order currently occupies, plus an extra command that tells the agent to wait where it is. In this case, a command is considered to be successfully completed by the agent if it passes through the specified door (or stays in the room, if commanded to wait), whilst ensuring that the room is cleared of enemies. The command is considered to have failed if the agent leaves through another door or takes too long to complete the command. At each timestep if no agents have completed/failed their order then no new orders are issued, however if any new orders are issued then all agents that were currently following the ``wait" order are given new orders too. This means that the commander is making decisions at a different timescale from the individual agents as well as at a higher level of granularity (the commander does not need to know about the detailed room layout, for example).

We also consider a variant of these settings in which the commander can induce more complex behaviour by effectively synchronising the agent's behaviour. With the base set of orders, the commander has no direct control over when the commands are executed --- i.e. if it tells an agent to go through a door, there is no mechanism for saying how quickly that order should be carried out. This means that in general it would not be able to learn to reliably send two agents through the same door at the same time. In the ``order-sync" variant of this set-up, we alter the meanings of the orders so that by default each order means go to the door but not through it --- unless the agent is already by the door in which case it means go through it. This means, for example, that the commander can get an agent to go to a door and then tell it to wait until the other agents are in place before going through. This introduces the potential for developing scenarios that involve more complicated coordination between agents and where the commander can learn more sophisticated behaviour.

The low-level agents have six actions available to them at each time step: wait, shoot or move north, south, east or west, and their aim is simply to learn how to complete the commands they are given. Combat between agents is kept simple, with line-of-sight calculations checking whether agents and enemy combatants can see each other and if so allowing a shoot action that does a fixed amount of damage.

\section{Methodology}
\subsection{Preliminaries}
RL takes place within the framework of Markov decision processes (MDPs). In general, we consider a system that can be in a set of possible states $\mathcal{S}$ and denote a given state at time step t as $s_t \in \mathcal{S}$, along with a set of possible actions it can take, $\mathcal{A}$ (denoting the action taken at time step t as $a_t \in \mathcal{A})$. A Markov decision process is then defined in terms of an initial distribution over starting states, $p(s_0)$, a transition function, $P(s_{t+1} | s_{t}, a_{t})$ (that depends only on the current state and the action taken) and a reward signal $r_{t} = R(s_{t+1}, s_t, a_t)$. The reinforcement learning problem is then to learn a policy over actions, $\pi^*(a_t | s_t)$, which maximises the discounted sum of rewards over an episode of length $T$, i.e.

\begin{equation}
    \pi^* = \argmax_{\pi} \ \mathbb{E}_{\subalign{s_0 & \sim p(s_0) \\ a_t & \sim \pi(s_t | a_t) \\ s_{t+1} & \sim P(s_{t+1} | s_t, a_t)}} \left[ \sum_{t=0}^{T} \gamma^t r_t \right]
\end{equation}

where $\gamma \in [0,1]$ is a discount factor that allows for more emphasis to be placed on immediate rewards over future rewards. In our Feudal HRL set up, we consider separate Markov chains for the commander and each of the low-level agents, denoting the states/actions/rewards of each as $(s_t^c, a_t^c, r_t^c)$ and $(s_t^a, a_t^a, r_t^a)$ respectively. Only the commander receives reward directly from the environment itself, and the agent's reward is entirely defined by the commander.

\subsection{Q-learning and Deep Q-Networks}
To initially test our environment and demonstrate the advantages of feudal HRL we start with a simple tabular Q-learning approach \cite{sutton_and_barto}, before going on to implement a more powerful approach based on deep Q-networks \cite{dqn}. In the tabular case we learn a ``lookup table" $Q(s,a)$ that estimates the value (i.e. expected future return) of the action $a$ from the state $s$. This can be used to induce a deterministic optimal policy:
\begin{equation}
\pi^* (a | s) = \begin{cases}1 \ \text{if} \ a = \argmax_{a' \in \mathcal{A}} Q(s, a') \\ 0 \ \text{otherwise}
\end{cases}
\end{equation}
i.e. choosing the action that leads to the maximum expected return. Since Q-learning is an off-policy algorithm, we can update this table using experience gathered with any given behavioural policy, $\pi_b(a | s)$. In practice we use an $\epsilon$-greedy version of the current optimal policy estimate to aid with exploration, where we choose a threshold $\epsilon \in [0, 1]$ and select random actions with probability $\epsilon$ and ``greedy" actions from the optimal policy with probability $(1 - \epsilon)$. Given a tuple of experience $(s, a, r, s', d)$, where $s'$ represents the next state and $d$ represents a boolean indicating if the episode has finished, the lookup table can be updated as follows:
\begin{equation}
\begin{split}
    Q(s, a) & = Q(s, a) \ + \\  & \alpha \left( Q(s, a) - (r + \gamma \ (1 - d) \max_{a'} Q(s', a')) \right) \ 
    \end{split}
\end{equation}
where $\alpha$ is a learning rate. The major advantages of tabular Q-learning is that generally the learning is both efficient and stable, however a key drawback is that it does not scale well to complex scenarios. This is because the lookup table has to store an entry for every possible state-action pair, $(s, a)$, and as the size of the state/action space grows this quickly becomes infeasible. A commonly used approach to scale Q-learning to larger state/action spaces is to use a high-capacity, deep neural network as a function approximator for the value function, $Q_{\theta}(s, a)$ (where $\theta$ represent the parameters of the neural network). In recent years this ``deep Q-network" approach has been widely applied with significant success in a variety of domains, although a couple of ``tricks" are required in order to stabilise training. Firstly, to prevent overfitting to the most recent experience only a replay buffer $\mathcal{R}$ is used to store a large amount of collected experience, and then batches are sampled randomly from this when training the network. The second trick to improve training stability is to use a separate target network $Q_{\tilde{\theta}}(s, a)$ which is updated much more slowly than the main value network, $Q_{\theta}(s, a)$, when calculating the target value. The value network is then trained using stochastic gradient descent on the following loss function:

\begin{equation}
\label{dqnloss}
\begin{split}
\mathbb{E}_{(s, a, r, s', d) \sim \mathcal{R}} \left[\left(Q_{\theta}(s, a) - (r + \gamma \ (1 - d) \max_{a'} Q_{\tilde{\theta}}(s', a')) \right)^2 \right]
\end{split}
\end{equation}

For our experiments we make use of another trick that has been shown to improve stability further and reduce the risk of learning overoptimistic value estimates --- ``Double Deep Q-Networks" (DDQN) \cite{ddqn}. This alters the value function target in equation \ref{dqnloss} from $r + \gamma \ (1-d) \max_{a'} Q_{\tilde{\theta}}(s',a')$ to $r + \gamma \ (1-d) Q_{\tilde{\theta}}(s', \argmax_{a'} Q_{\theta}(s', a'))$, representing a minor algorithmic change.

\subsection{Feudal HRL applied to Room Clearance}
In our set-up the process of learning to solve a given room clearance scenario is essentially broken down into two different problems --- training the commander and training the agents. To keep things simple, we model all of the agents as identical. In the tabular case this means the agents share the same lookup table, whereas with the DDQN approach they all share a single action-value neural network, $Q^a_{\theta}(s, a)$, which is trained from a shared replay buffer containing experience gathered by all agents. We condition the agent's observation with the current command it has been given, meaning in effect we are training a goal-conditioned policy for the agents. From the agent's perspective, an ``episode" begins when they are first given an order and ends either when they have completed/failed that order or when they run out of time (at which point the done boolean, $d$, will be set to true, and a new agent episode will begin upon them receiving their next order). These can be thought of as much shorter ``sub-episodes" of a full room clearance episode, meaning that the agents only have to focus on significantly simpler tasks over a much shorter period of time than the commander. The rewards the agent receives are defined by the commander rather than the underlying reward provided by the environment, and for simplicity we choose these to be:

\begin{equation}
    r_t^a(s_t^a, a_t^a, s_{t+1}^a) = \begin{cases}
    +10 \ &\text{if $s_{t+1}^a$ completes given order} \\
    -10 \ &\text{if $s_{t+1}^a$ completes wrong order} \\
    \ \ 0 \ &\text{otherwise}
    \end{cases}
\end{equation}

The agent's state representation only includes local information about the room it is in --- since global information about the building layout is irrelevant for learning how to complete the order it has been given. The state consists of a flattened 2D array of the room which identifies walkable cells, as well as the coordinates of each of the doors within the current room and the current order the agent is to follow.

The commander's MDP is slightly more complicated because it only makes decisions when an agent requires a new order, which sometimes means it has to wait multiple timesteps in the physical environment before making a decision, whilst at other times is required to make multiple decisions in a single timestep. Each time it has to make a decision, we provide it with a state $s_t^c$ which gives it information about the current global room layout (which rooms are clear, which are unclear), information about the current agent it is giving an order to (which room it occupies, distances to doors within the room etc) and information about where the other agents are and which orders they are currently following. The reward it receives is the sum of environment rewards given in the time interval between decisions (or zero between two decisions made on the same time step).

In the primary implementation, we keep a separate replay buffer for the commander's experience and train a separate DDQN value function (or lookup table in the tabular case). To try and ensure that the effect of different timescales of the two learning processes is minimised we perform the training updates separately but at the same effective rate --- so a training step with the agent is performed every time a new piece of experience is added to the agent's buffer, and a training step with the commander is performed every time the commander's buffer is added to. For more complicated scenarios, we also consider experiments where we pre-train the agents to learn to complete orders before training the commander (alternatively, we also consider scripting the behaviour of the agents for some scenarios). The reason for this is that whilst the agents are still learning to complete orders the learning process for the commander is effectively non-stationary --- since its orders might be followed or might not, depending on the learning progress of the agents. This makes it difficult for the commander to learn an accurate value of a given state because this will be effected by subsequent changes in behaviour of the learning agents. However, once the agents have learned to succeed at completing their orders (or if their behaviour is scripted), we then start to see huge benefits from feudal HRL over standard multi-agent algorithms. This is because at that point from the commander's point of view the MDP is completely stationary, i.e. the environment (including the agents) behave in a completely consistent way.

The environment reward function is something which can vary depending on the scenario (and something we investigate in sections V(C) and V(E)), but we consider the ``default" reward function to be the following:
\begin{equation}
\label{commander_reward_fn}
    r_t^c(s_t^c, a_t^c, s_{t+1}^c) = \begin{cases}
    R_{\text{complete}} \ \text{if building is clear} \\
    -1 + \frac{\text{currently clear rooms}}{\text{total rooms}} \ \text{otherwise}
    \end{cases}
\end{equation}
where $R_{\text{complete}}$ is a large positive reward given at the end when the building is clear. The negative reward at each step encourages the building to be cleared as quickly as possible.

\section{Results}
\subsection{Tabular Feudal HRL}
\begin{figure}
    \centering
    \includegraphics[width=1.0\linewidth]{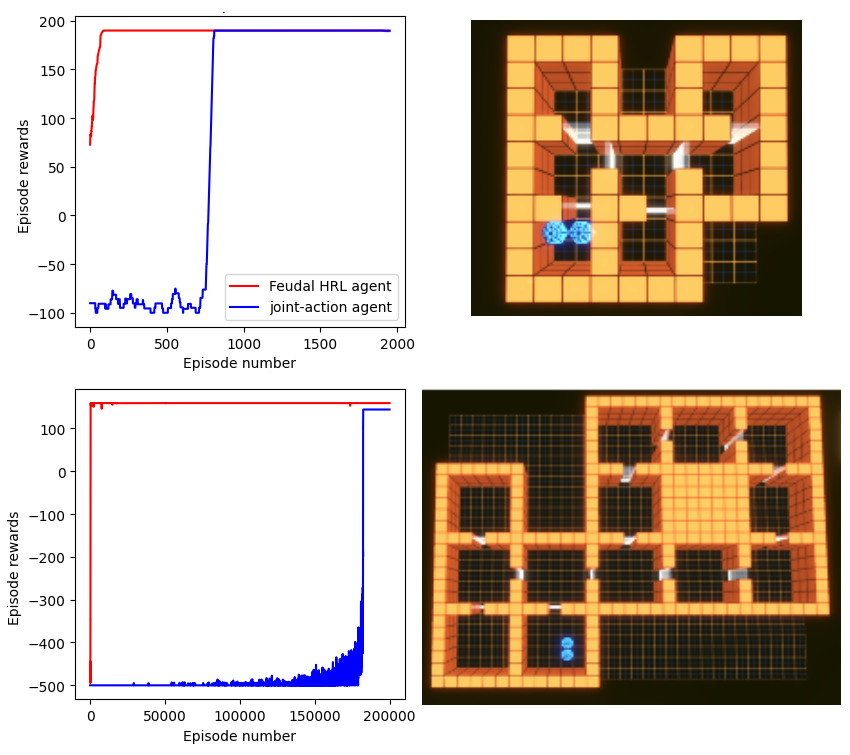}
    \caption{Tabular RL setting - Feudal HRL agent learns significantly more quickly than a naive joint-action learner.}
    \label{fig:tabular_rl}
\end{figure}

\begin{figure*}
    \centering
    \includegraphics[width=0.8\linewidth]{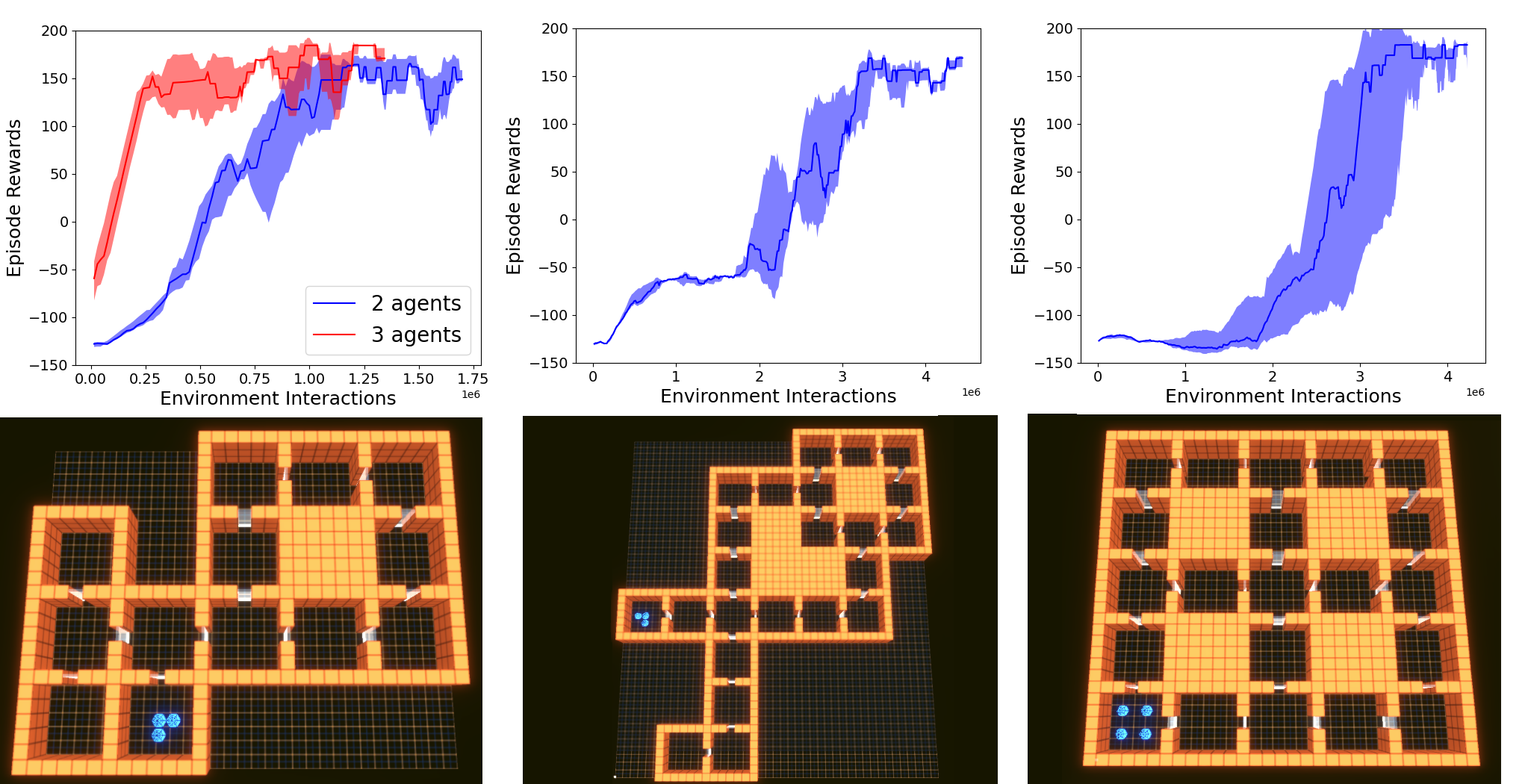}
    \caption{Performance of our Feudal DDQN agent on three maps of increasing complexity}
    \label{fig:ddqn_hrl_res}
\end{figure*}
We start by testing a simple tabular implementation of Feudal HRL and apply it to two relatively simple maps, shown in Figure \ref{fig:tabular_rl}. For both of these we consider only two agents and a commander, since we find that the tabular approach does not scale well to more agents (the number of possible states for the commander increases exponentially with the number of agents). As a simple demonstration of the utility of the Feudal HRL framework for these problems, we test against a naive ``joint-action agent". The joint-action agent considers the whole process as a single MDP and at each step feeds an action which consists of a concatenation of all of the actions for each individual agent. So for example, if each agent can go North, East, South or West and there are two agents, the joint-action learner would have 16 actions available to it (North-North, North-East, etc). In the two experiments we run here we use a completely sparse reward (i.e. only give a large positive reward once the building is clear), rather than the denser reward described in equation \ref{commander_reward_fn}. Unsurprisingly, we find that the Feudal HRL approach enormously increases the speed of learning. This is particularly striking for the larger map, where the Feudal HRL agent solves the task in about 300 episodes, whereas the joint-action learner takes over 150,000 (and still converges to a slightly sub-optimal solution).

\subsection{Feudal HRL with DDQNs}

As mentioned in Section IV, tabular Q-learning struggles as the state space starts to grow since every distinct state-action pair has to have a value stored in a lookup table. Looking at the larger map in Figure \ref{fig:tabular_rl}, our feudal tabular Q-learning algorithm is unable to solve this when the commander is in control of three agents instead of two since the number of unique states it can observe is simply too large. In this section we demonstrate how our Feudal DDQN approach readily scales to larger, more complicated maps involving a larger number of agents.

Firstly, we consider the larger map from Figure \ref{fig:tabular_rl} with both two and three agents. Whilst the tabular approach did not work here with three agents, in Figure \ref{fig:ddqn_hrl_res} we see that the DDQN approach learns to clear the building more quickly with three agents than with two (this makes sense as the logic of clearing the building is actually simpler with three agents). The second room layout consists of four corridors and a loop, and can only be solved with three agents (it is also significantly larger than the first map). The third map involves four intersecting loops, and requires the careful coordination of four agents in order to solve it (it is very easy to unclear multiple rooms with a single misstep). All of the plots show the median performance from initialising the training process with three different random seeds, with the shaded area representing the standard deviation. We see that for each of these maps our approach is consistently able to learn to solve them, with fairly stable training progress.

\subsection{Command Synchronisation}
\begin{figure}
    \centering
    \includegraphics[width=0.8\linewidth]{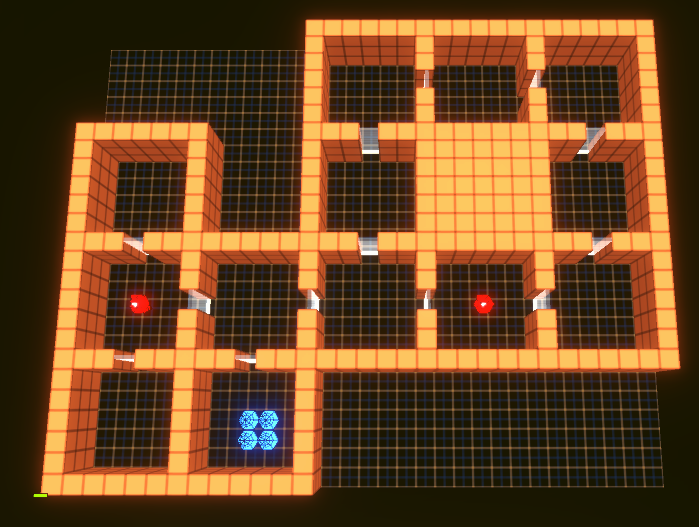}
    \caption{Scenario explored with the command synchonisation variant of our set-up. Rooms with enemies in can only be cleared without casualty if the commander learns to send two agents into the room at the same time.}
    \label{fig:commandsync}
\end{figure}
We now consider the described variant where the commander is able to sync the agents' behaviour by telling them to stand/wait by doors before entering them. In this case we have to modify the commander's state to give it information about which doors have agents waiting by them as well as if other agents have been ordered to go through them. The scenario we look at is a simple map with two enemies. We set it up so that both agents and enemies require two shots to kill, which means that if an agent goes into a room by itself it can kill the enemy if it shoots twice but it will also die. However, if two agents are sent into the room at the same time it is possible to kill the enemy without losing any agents. The map we consider is shown in Figure \ref{fig:commandsync}. If we train using the default reward function described in equation \ref{commander_reward_fn} then the commander is indifferent as to whether or not the agents survive and only cares about clearing the building as quickly as possible. As such, it learns a policy in which usually at least one agent dies. On the other hand, if we modify the reward slightly and introduce a negative reward for the commander each time an agent dies then it is always able to learn to keep all four agents alive. [include videos in SI which we can refer to for comparison].

\subsection{Scaling to Realistic Floorplans}
\begin{figure}
    \centering
    \includegraphics[width=1.0\linewidth]{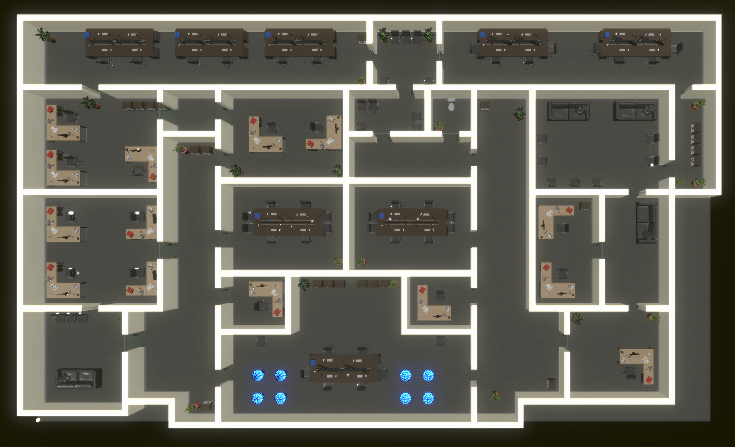}
    \caption{A more realistic floor-plan.}
    \label{fig:realistic_floorplan}
\end{figure}

In this section we look at scaling what we have to more realistic building floorplans (see e.g. Figure \ref{fig:realistic_floorplan}). The big problem that arises is that in situations where we have many non-uniform rooms training the agents to complete their orders is significantly more difficult. This is particularly true when it takes a long time for the commander to learn how to reach certain rooms, such that for those cases the agents receive little or no example data to train on. This means it can take a long time for the lower-level's agent policy to converge, making the environment from the commander's point of view appear non-stationary for a much longer period of time. This inconsistency in terms of whether or not an agent successfully completes an order then also makes it more challenging to train the commander, since it is more difficult to determine if an order it gives is actually good/bad or whether an agent just didn't actually do what it was told to. As such, solving the map shown in Figure \ref{fig:realistic_floorplan} with 8 agents is not possible with the approach described so far, and in fact almost no progress is made when we try to do so.

We look at two approaches for getting around this. Firstly, we consider training the agent's policies separately and before we start training the commander at all. To do this, we initialize agents randomly in different rooms and give them random orders to complete. This allows the agents to gather relevant experience from all of the rooms in similar proportions, making it substantially easier to train their policy. Once we have finished training the agent policy we then freeze it and train the commander, who now is in control of agents that all have completely stationary policies. The downside of this is that it is a somewhat less general approach --- it requires that we modify the environment slightly to allow the agents to be initialised in random starting rooms. However, it is easy to imagine other environments where it would be possible to train the ``lower-level" agents to complete their orders separately before training the commander, and from a pragmatic point of view it offers considerable advantages.

An even more pragmatic approach here is to simply script the behaviour of the low-level agents, since the behaviour they learn is fairly simple (go to a given door whilst making sure to clear the room of enemies first, if applicable). Ultimately for our setup (where the agents' behaviour is particularly simple) this leads to the same situation as pre-training the agent's policy, but saves time as no training is required for this step. Obviously this is even less general than the pre-training approach, however in situations where it is possible it also offers obvious benefits. Furthermore, it is easy to imagine applications of RL in more advanced simulations where it would be possible to script the behaviour of individual agents to perform their comparatively simple orders whilst letting the RL focus purely on the higher-level strategy required by a commander. Both of these approaches allow us to solve the map shown in Figure \ref{fig:realistic_floorplan}.

\subsection{Moving Gunman and Civilian Rescue Scenario}
\begin{figure}
    \centering
    \includegraphics[width=1.0\linewidth]{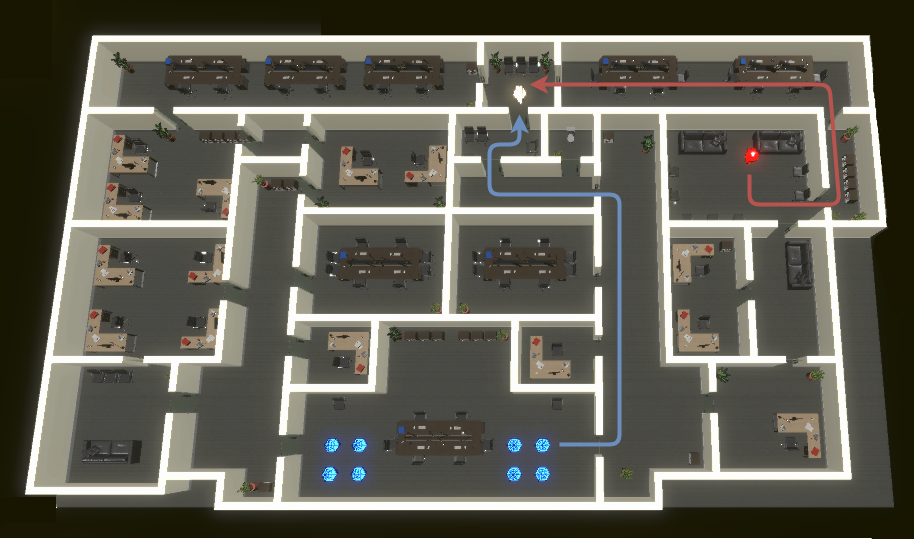}
    \caption{Scenario within the office floor-plan. A moving enemy makes its way towards a stationary civilian, and opens fire if not intercepted in time by one of the blue agents.}
    \label{fig:roaming_gunman}
\end{figure}

To introduce a bit more complexity into the set-up we consider the floor-plan from Figure \ref{fig:realistic_floorplan} and introduce a moving enemy and a stationary civilian (see Figure \ref{fig:roaming_gunman}). The enemy deterministically follows a path towards the room with the civilian and if it gets there before any of the agents do will open fire. The set-up is such that it is possible for an agent to get to the room in time to intercept the enemy before this happens, but only just. Whilst fairly contrived, this allows us to demonstrate how providing an RL agent with different objectives can lead to different emergent behaviours.

We consider comparing the behaviour with the default reward function in equation \ref{commander_reward_fn} with the following reward function that places importance on ensuring no civilian casualties:

\begin{equation}
\label{civilian_rew_fn}
    r_t^c(s_t^c, a_t^c, s_{t+1}^c) = \begin{cases}
    R_{\text{complete}} \ \text{if building is clear} \\ \ \ \ \ \ \ \ \ \ \ \ \ \ \text{and civilian is alive} \\
    - R_{\text{complete}} \ \text{if civilian dies} \\
    -1 + \frac{\text{currently clear rooms}}{\text{total rooms}} \ \text{otherwise}
    \end{cases}
\end{equation}

With the default reward function the commander is only interested in clearing the building as quickly as possible, which does not involve rushing to the room containing the civilian. As such, it learns a policy that clears the building efficiently but which allows the civilian to die. On the other hand, with the modified reward function there is a clear incentive to ensure the civilian survives, and so the commander learns to immediately rush an agent to the room containing the civilian and intercept the enemy (at the cost of slowing down the overall building clearance time). Whilst this is a somewhat contrived example, it demonstrates how we can build different preferences into a reward function which then automatically leads to the system learning qualitatively different strategies. 

\section{Conclusions}
We introduced a flexible and customisable tool that allows us to build simple maps with multiple agents, enemies and civilians. Whilst it is possible to build any ruleset on top of these maps, we focused our experiments on room clearance. We showed that on relatively simple maps an algorithm based on Feudal Hierarchical RL was able to learn solutions that required the coordination of multiple agents significantly more efficiently than other multi-agent RL algorithms. We then went on to show how this could be extended to solving more realistic floorplans with a larger number of agents, either by pre-training the lower-level agents or directly scripting their behaviour, and investigated how the commander can learn different strategies based on specified preferences built into the reward function.

This can be considered as an important initial step towards finding useful applications of RL within the domains of military concept analysis and wargaming, which often involve many agents and where a command hierarchy is already in place. In the future we also hope that this tool can be useful in developing further research into multi-agent algorithms that can scale solve a more challenging array of problems.
\section*{Acknowledgment}
The authors are extremely grateful to the Defence Science and Technology Laboratory for funding this work.
\bibliographystyle{IEEEtran}
\bibliography{IEEEabrv,references}

\begin{thebibliography}{10}
\providecommand{\url}[1]{#1}
\csname url@samestyle\endcsname
\providecommand{\newblock}{\relax}
\providecommand{\bibinfo}[2]{#2}
\providecommand{\BIBentrySTDinterwordspacing}{\spaceskip=0pt\relax}
\providecommand{\BIBentryALTinterwordstretchfactor}{4}
\providecommand{\BIBentryALTinterwordspacing}{\spaceskip=\fontdimen2\font plus
\BIBentryALTinterwordstretchfactor\fontdimen3\font minus
  \fontdimen4\font\relax}
\providecommand{\BIBforeignlanguage}[2]{{%
\expandafter\ifx\csname l@#1\endcsname\relax
\typeout{** WARNING: IEEEtran.bst: No hyphenation pattern has been}%
\typeout{** loaded for the language `#1'. Using the pattern for}%
\typeout{** the default language instead.}%
\else
\language=\csname l@#1\endcsname
\fi
#2}}
\providecommand{\BIBdecl}{\relax}
\BIBdecl

\bibitem{alphazero}
\BIBentryALTinterwordspacing
D.~Silver, T.~Hubert, J.~Schrittwieser, I.~Antonoglou, M.~Lai, A.~Guez,
  M.~Lanctot, L.~Sifre, D.~Kumaran, T.~Graepel, T.~P. Lillicrap, K.~Simonyan,
  and D.~Hassabis, ``Mastering chess and shogi by self-play with a general
  reinforcement learning algorithm,'' \emph{CoRR}, vol. abs/1712.01815, 2017.
  [Online]. Available: \url{http://arxiv.org/abs/1712.01815}
\BIBentrySTDinterwordspacing

\bibitem{alphago}
\BIBentryALTinterwordspacing
D.~Silver, A.~Huang, C.~J. Maddison, A.~Guez, L.~Sifre, G.~van~den Driessche,
  J.~Schrittwieser, I.~Antonoglou, V.~Panneershelvam, M.~Lanctot, S.~Dieleman,
  D.~Grewe, J.~Nham, N.~Kalchbrenner, I.~Sutskever, T.~Lillicrap, M.~Leach,
  K.~Kavukcuoglu, T.~Graepel, and D.~Hassabis, ``Mastering the game of go with
  deep neural networks and tree search,'' \emph{Nature}, vol. 529, no. 7587,
  pp. 484--489, Jan 2016. [Online]. Available:
  \url{https://doi.org/10.1038/nature16961}
\BIBentrySTDinterwordspacing

\bibitem{alphastarblog}
O.~Vinyals, I.~Babuschkin, J.~Chung, M.~Mathieu, M.~Jaderberg, W.~Czarnecki,
  A.~Dudzik, A.~Huang, P.~Georgiev, R.~Powell, T.~Ewalds, D.~Horgan, M.~Kroiss,
  I.~Danihelka, J.~Agapiou, J.~Oh, V.~Dalibard, D.~Choi, L.~Sifre, Y.~Sulsky,
  S.~Vezhnevets, J.~Molloy, T.~Cai, D.~Budden, T.~Paine, C.~Gulcehre, Z.~Wang,
  T.~Pfaff, T.~Pohlen, D.~Yogatama, J.~Cohen, K.~McKinney, O.~Smith, T.~Schaul,
  T.~Lillicrap, C.~Apps, K.~Kavukcuoglu, D.~Hassabis, and D.~Silver,
  ``{AlphaStar: Mastering the Real-Time Strategy Game StarCraft II},''
  \url{https://deepmind.com/blog/alphastar-mastering-real-time-strategy-game-starcraft-ii/},
  2019.

\bibitem{openai2019dota}
\BIBentryALTinterwordspacing
OpenAI, C.~Berner, G.~Brockman, B.~Chan, V.~Cheung, P.~Dębiak, C.~Dennison,
  D.~Farhi, Q.~Fischer, S.~Hashme, C.~Hesse, R.~Józefowicz, S.~Gray,
  C.~Olsson, J.~Pachocki, M.~Petrov, H.~P. de~Oliveira~Pinto, J.~Raiman,
  T.~Salimans, J.~Schlatter, J.~Schneider, S.~Sidor, I.~Sutskever, J.~Tang,
  F.~Wolski, and S.~Zhang, ``Dota 2 with large scale deep reinforcement
  learning,'' 2019. [Online]. Available: \url{https://arxiv.org/abs/1912.06680}
\BIBentrySTDinterwordspacing

\bibitem{visuomotor}
\BIBentryALTinterwordspacing
S.~Levine, C.~Finn, T.~Darrell, and P.~Abbeel, ``End-to-end training of deep
  visuomotor policies,'' \emph{CoRR}, vol. abs/1504.00702, 2015. [Online].
  Available: \url{http://arxiv.org/abs/1504.00702}
\BIBentrySTDinterwordspacing

\bibitem{openai2021asymmetric}
O.~OpenAI, M.~Plappert, R.~Sampedro, T.~Xu, I.~Akkaya, V.~Kosaraju,
  P.~Welinder, R.~D'Sa, A.~Petron, H.~P. de~Oliveira~Pinto, A.~Paino, H.~Noh,
  L.~Weng, Q.~Yuan, C.~Chu, and W.~Zaremba, ``Asymmetric self-play for
  automatic goal discovery in robotic manipulation,'' 2021.

\bibitem{HRL}
\BIBentryALTinterwordspacing
A.~G. Barto and S.~Mahadevan, ``Recent advances in hierarchical reinforcement
  learning,'' \emph{Discrete Event Dynamic Systems}, vol.~13, no. 1–2, p.
  41–77, Jan. 2003. [Online]. Available:
  \url{https://doi.org/10.1023/A:1022140919877}
\BIBentrySTDinterwordspacing

\bibitem{feudalrl92}
\BIBentryALTinterwordspacing
P.~Dayan and G.~E. Hinton, ``Feudal reinforcement learning,'' in \emph{Advances
  in Neural Information Processing Systems}, S.~Hanson, J.~Cowan, and C.~Giles,
  Eds., vol.~5.\hskip 1em plus 0.5em minus 0.4em\relax Morgan-Kaufmann, 1993.
  [Online]. Available:
  \url{https://proceedings.neurips.cc/paper/1992/file/d14220ee66aeec73c49038385428ec4c-Paper.pdf}
\BIBentrySTDinterwordspacing

\bibitem{vezhnevets2017feudal}
A.~S. Vezhnevets, S.~Osindero, T.~Schaul, N.~Heess, M.~Jaderberg, D.~Silver,
  and K.~Kavukcuoglu, ``Feudal networks for hierarchical reinforcement
  learning,'' 2017.

\bibitem{dqn}
V.~Mnih, K.~Kavukcuoglu, D.~Silver, A.~Graves, I.~Antonoglou, D.~Wierstra, and
  M.~Riedmiller, ``Playing atari with deep reinforcement learning,'' 2013.

\bibitem{ahilan2019feudal}
S.~Ahilan and P.~Dayan, ``Feudal multi-agent hierarchies for cooperative
  reinforcement learning,'' 2019.

\bibitem{sutton_and_barto}
\BIBentryALTinterwordspacing
R.~S. Sutton and A.~G. Barto, \emph{Reinforcement Learning: An
  Introduction}.\hskip 1em plus 0.5em minus 0.4em\relax Cambridge, MA, USA: MIT
  Press, 1998. [Online]. Available:
  \url{http://www.cs.ualberta.ca/%7Esutton/book/ebook/the-book.html}
\BIBentrySTDinterwordspacing

\bibitem{ddqn}
H.~van Hasselt, A.~Guez, and D.~Silver, ``Deep reinforcement learning with
  double q-learning,'' 2015.

\end{thebibliography}

\end{document}